\pdfoutput=1

\documentclass[11pt]{article}
\usepackage[dvipsnames]{xcolor}
\usepackage{caption}
\usepackage{subfig}

\usepackage[final]{acl}

\usepackage{times}
\usepackage{latexsym}
\usepackage{graphicx}
\usepackage{array, tabularx, boldline}
\usepackage{cellspace}
\usepackage{microtype}
\usepackage{inconsolata}
\usepackage{multirow}
\usepackage{makecell}
\usepackage[T1]{fontenc}

\usepackage[utf8]{inputenc}

\usepackage{microtype}

\usepackage{inconsolata}

\title{Two-Pronged Human Evaluation of ChatGPT Self-Correction in Radiology Report Simplification
}

\author{Ziyu Yang \\ CIS, Temple University \\ \texttt{zyyang@temple.edu}
        \And 
        Santhosh Cherian \\ Temple University Hospital \\ \texttt{santhosh.cherian@tuhs.temple.edu} 
        \AND
        Slobodan Vucetic \\ CIS, Temple University \\ \texttt{vucetic@temple.edu}}

\begin{document}
\maketitle
\begin{abstract}
Radiology reports are highly technical documents aimed primarily at doctor-doctor communication. There has been an increasing interest in sharing those reports with patients, necessitating providing them patient-friendly simplifications of the original reports. This study explores the suitability of large language models in automatically generating those simplifications. We examine the usefulness of chain-of-thought and self-correction prompting mechanisms in this domain. We also propose a new evaluation protocol that employs radiologists and laypeople, where radiologists verify the factual correctness of simplifications, and laypeople assess simplicity and comprehension. Our experimental results demonstrate the effectiveness of self-correction prompting in producing high-quality simplifications. Our findings illuminate the preferences of radiologists and laypeople regarding text simplification, informing future research on this topic.
\end{abstract}

\section{Introduction}

An increasing number of healthcare providers are interested in sharing  health records with patients. That is a positive development because research has shown that sharing medical records with patients might improve patient-doctor communication~\cite{ross2003effects}, increase patient involvement in care~\cite{delbanco2012inviting}, and  improve outcomes~\cite{rosenkrantz2015survey}. However, the health literacy~\cite{kutner2006health} of most patients is often not sufficient to enable an understanding of their health records \cite{lalor2018comprehenotes}. The health literacy gap is especially severe for some types of medical reports, such as radiology reports, whose primary purpose is doctor-doctor communication. As a result, radiology reports use particularly complex medical jargon and highly specialized descriptions \cite{delbanco2012inviting} and present a particular challenge for patients \cite{hong2017supporting}. For instance, a recent study \cite{yi2019readability} found that the mean readability grade level of MRI reports was above the 12th-grade reading level. Without adequate counseling with an experienced clinician, the severity of the radiology findings may be misinterpreted by the patients. It could lead to unnecessary stress, improper follow-up, and even to increased patient mortality \cite{sudore2006limited}. 

There is an increasing interest in patient-friendly reporting. One way to accomplish this is to ask radiologists to supplement their expert-language reports with patient-friendly summaries. One downside of this approach is a negative impact on radiologists' cognitive load and productivity. Another downside is the curse of knowledge \cite{camerer1989curse}, making it challenging for radiologists to simplify their reports. An enticing alternative that has garnered much recent interest \cite{jeblick2022chatgpt, lyu2023translating} is to generate patient-friendly simplifications with large language models (LLMs) and ask  radiologists to check the generated simplifications before releasing them.

There are several open challenges to the generation of patient-friendly radiology reports. The first is that it needs to be clarified what constitutes a good simplification. The existing research has varying views of the trade-offs between factuality, completeness, simplicity, and brevity \cite{jiang2020neural, cripwell2022controllable}. A proper combination of these measures may depend on individual user preferences. As a result, it would be very challenging to create a widely acceptable parallel corpus for radiology report simplification. Moreover, very different simplifications could be evaluated as equally successful (longer and more detailed versus shorter with only critical information). Thus, even if the parallel corpus was created and used to train and test an LLM, automatic evaluation using measures that rely on sequence similarity \cite{lin2004rouge, xu2016optimizing, zhang2019bertscore} might be misleading. Instead, until there is more clarity about what constitutes a reasonable radiology simplification, we think humans should perform the evaluation. 

There is no broadly accepted protocol for human evaluation of simplified expert text \cite{van2019evaluating, devaraj2022evaluating, lu2023napss}, including what questions to ask and who should answer them. In this paper, we propose a novel evaluation protocol following two ideas. First, we observe that laypeople should not be asked factuality and completeness questions due to the lack of expert knowledge and that radiologists should not be asked about simplicity due to the curse of knowledge bias. Thus, our protocol employs laypeople and radiologists with slightly different questions. Second, we observe that a good simplification is the one that increases understanding compared to the original text, but also that there can be a dangerous mismatch between perceived and actual understanding. Thus, laypeople are asked both about their perception and their actual increase in understanding when an expert text is supplemented by its simplification.   

Another contribution of this paper is in evaluating the capabilities of the state-of-the-art LLMs without constructing a large parallel text corpus. Arguably, the best publicly available LLM at the moment is ChatGPT  \cite{openai2023gpt4}, and recent papers \cite{jeblick2022chatgpt, lyu2023translating} indicate that both its 3.5 and 4 versions can provide high-quality radiology report simplifications only through prompting. In this paper, we provide an in-depth evaluation of chain-of-thought (CoT) prompting \cite{wei2022chain} and self-correction \cite{madaan2023self}. In the CoT approach, LLMs are prompted to justify an answer before providing the answer. In the self-correction approach, LLMs are prompted to critique their original response and asked to consider the critique to give an improved response. Both methods have been shown to work well in several applications \cite{fu2023improving, chen2023teaching}. To our knowledge, they have yet to be evaluated on radiology report simplification.


We designed experiments to answer the following research questions: ($\mathcal{Q}_1$) Is the proposed human evaluation protocol insightful? ($\mathcal{Q}_2$) Are CoT and self-correction helpful in the simplification of radiology reports? ($\mathcal{Q}_3$) What is the relationship between perceived and actual understanding of radiology reports? ($\mathcal{Q}_4$) What kinds of simplifications are preferred by experts and laypeople? The answers should be informative for future research towards high-quality simplifications of expert texts. 

\section{Related Work}

\subsection{Radiology text simplification}
Traditionally, text simplification referred to lexical simplification that paraphrases text \cite{chen2018natural, biran2011putting, weng2018mapping}. More recently, it shifted towards semantic simplification that seeks to simplify grammatically complex text \cite{shardlow2014survey, leroy2016effects}. This paper adopts this more novel emphasis. Plain language summarization \cite{guo2021automated, devaraj2021paragraph} is an alternative term that reminds that the main objective is to enhance laypeople's understanding of expert-written texts. We think that summarization is not the best term because it implies text compression, while text simplification is more interested in text understanding, which allows creation of text longer than the original.

Radiology text simplification using LLMs has recently drawn significant attention \cite{ondov2022survey}. A recent work used fine-tuned BART \cite{lewis2019bart} to simplify 140 liver-related radiology sentences \cite{yang2023data}.  In \cite{jeblick2022chatgpt, lyu2023translating}, researchers explored the use of prompt learning with the GPT family \cite{brown2020language}, including ChatGPT-3.5 and ChatGPT-4, to simplify radiology reports. \cite{jeblick2022chatgpt} focused on three artificial reports, while \cite{lyu2023translating} considered over 100 reports. However, they did not provide fully coherent evaluation of the simplifications.   

There are two related NLP problems that have been popular in radiology. Radiology report generation refers to automated creation of reports from X-ray or other radiographic images \cite{liu2023systematic}. This is an image-to-text task with a different set of objectives from text simplification.  Radiology report summarization refers to condensing the detailed "Findings" section of radiology reports into a succinct "Impression" section \cite{zhang2018learning}. This involves creating a shorter version of the report that retains all critical information without a necessity to make it clearer to laypeople \cite{chaves2022automatic, liang2022fine}. 

\subsection{Evaluation of text simplifications}
Assessing output of LLMs is integral to text simplification \cite{van2019evaluating, cripwell2022controllable} and related natural language generation tasks. 
In text simplification, automatic metrics such as ROUGE \cite{lin2004rouge} and BERTScore \cite{zhang2019bertscore} have been popular, which compare similarity between gold standard and generated sentences. SARI \cite{xu2016optimizing} compares simplified text both with reference simplifications and the original sentences, thus assessing the operation of adding, deleting, and keeping words. Unfortunately, these metrics often correlate poorly with human evaluation of text simplification \cite{alva2021suitability, liu2023gpteval, guo2023appls}. For readability assessment, the Flesch-Kincaid Grade Level (FKGL), Gunning Fog Index (GFI), and Automated Readability Index (ARI) are widely recognized metrics that estimate the text's reading difficulty. More recently, \cite{guo2023appls} proposed to assess readability by using difference in normalized perplexity scores from in-domain and out-of-domain language models. 


Using human evaluators has been increasingly popular in text simplification, despite the significant associated costs. Researchers typically evaluate fluency, adequacy, factuality, and simplicity of the simplified texts \cite{jiang2020neural, cripwell2022controllable}. Very often, these measures are vaguely defined and subject to interpretation. Recently, factuality was formalized in terms of addition, substitution, and deletion of information \cite{devaraj2022evaluating}.

In radiology report simplifications, there is no clear standard for evaluation. \cite{jeblick2022chatgpt} enlisted 15 radiologists to assess simplified reports for factual correctness, completeness, and potential harm. \cite{lyu2023translating} invited two radiologists to evaluate the simplified reports based on metrics such as information loss, misinterpretation, and an overall score. Interestingly, these studies did not evaluate the simplicity of the text.  \cite{lu2023napss} focused on simplicity, fluency, and factual accuracy. The study recruited students to assess factualness and simplicity and two medical experts to examine the factual consistency. However, it is unclear if the participating students possessed any medical expertise to represent laypeople and if they were qualified to assess the factualness.

\subsection{Prompting strategies for LLMs}
Modern LLMs can solve various NLP tasks with high success through prompting and without necessitating fine-tuning \cite{brown2020language}. The quality of output is very sensitive to prompting.  While prompting is sometimes considered an art form, there are a few strategies that work more often than not. One is Chain-of-thought (CoT) \cite{wei2022chain}. Another is self-correction \cite{chen2023teaching, madaan2023self}. \cite{huang2022large} self-improves an LLM through iterative fine-tuning. \cite{bai2022constitutional} leverages AI-generated feedback through reinforcement learning.  \cite{li2023camel} allows LLMs to self-improve their generations without training. They instantiate multiple LLMs models as different agents and let them collaborate towards better generation.

\section{Evaluation Protocol}
\label{eval}
As described in the previous section,  prior text simplification research used human evaluation, but did not clarify the roles of experts and laypeople in  evaluation. In the following, we will propose an evaluation protocol that  defines those roles.
\begin{figure*}[h]
  \includegraphics[width=\textwidth]{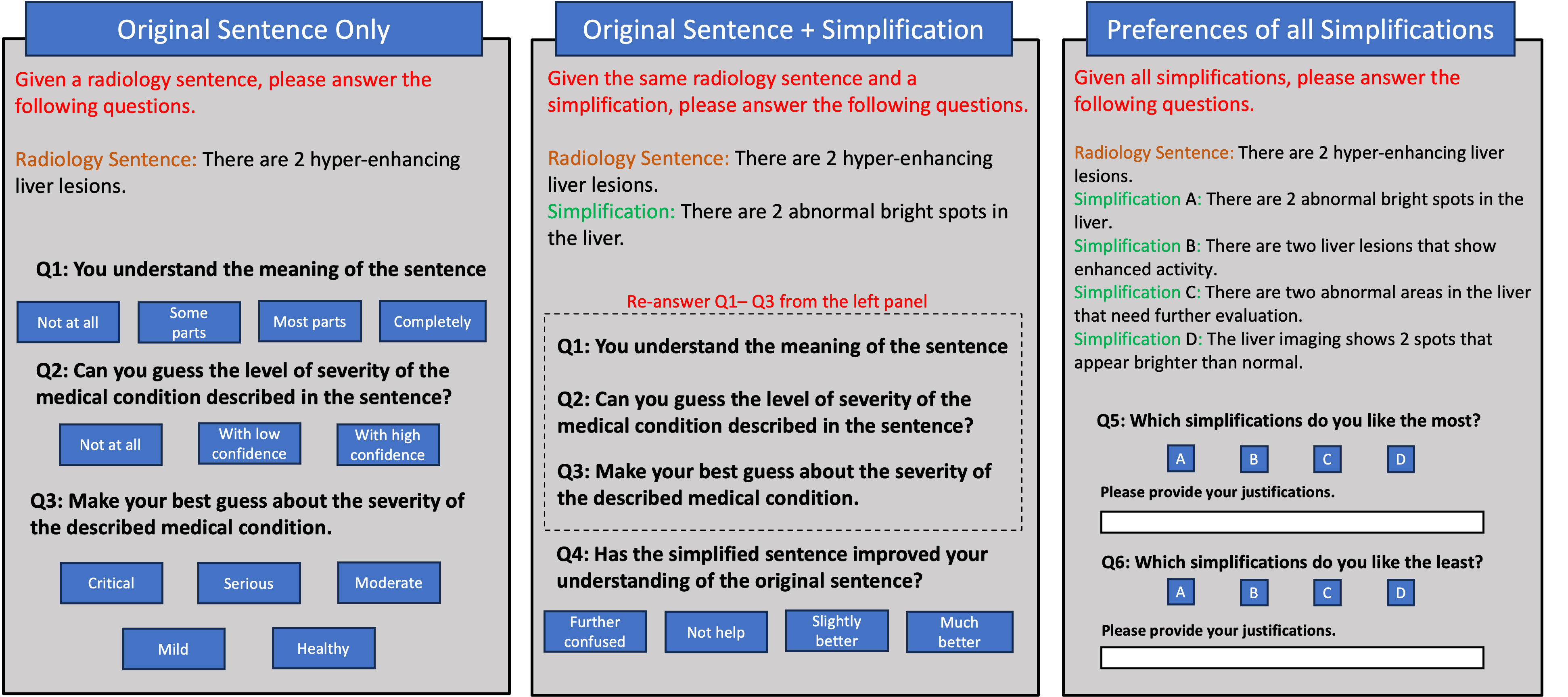}
  \caption{Layperson evaluation of radiology report simplifications. (a) (\textbf{left panel}) evaluates whether laypeople understand the original sentence. (b) (\textbf{middle panel}) evaluates whether simplification improves understanding. (c) (\textbf{right panel}) evaluates the preferences given a set of candidate simplifications and asks for justification.
}
  \label{fig:sur_lay}
\end{figure*}

\subsection{Factuality (for experts)}
\label{subsec:fact}
Factuality refers to correctly maintaining the original information. Motivated by \cite{devaraj2022evaluating} and \cite{jeblick2022chatgpt}, we measure three aspects of factuality.

\begin{itemize}
    \item \textbf{Correctness}: (Factualness/Substitution) Evaluates whether the simplification correctly interprets the information in the original sentence. 
    \item \textbf{Completeness}: (Adequacy/Meaning preservation/Deletion) Evaluates if there is any significant information loss in the simplification compared to the original text. Simplifications should retain all critical information from the original text, but it might be permissible to ignore less important information.
    \item \textbf{Hallucination}: (Addition) Evaluates if simplifications contain wrong statements or hallucinate new information that may misguide laypeople. 
\end{itemize}

We introduce a new measure that is related to Completeness.
\begin{itemize}
    \item \textbf{Structure}: Refers to a desire that simplifications follow a certain structure. Specifically, a good radiology simplification should mention: \textit{body parts, findings, and consequences}. 
    Body parts specify the anatomies and organs referred to in the radiology sentence (such as kidneys).
    Findings refer to the key observations in the radiology sentence (such as injuries or masses).
    Consequences refer to what findings indicate, which might not be explicitly stated in the original sentence, such as severity, certainty, and follow up. 
\end{itemize}

Only experts can adequately evaluate factuality and structure. Appendix \ref{subsec:rad} shows the exact survey design for expert evaluation of simplifications we used in our experiments. Note that we also ask the radiologists to evaluate the simplicity of generated simplifications for our analysis.

\subsection{Simplicity (for laypeople)}
\label{subsec:simp}
In prior work, simplicity mostly refers to readability, which measures text fluency and complexity of terms and grammar correctness. However, LLMs typically generate very fluent text, so evaluating that aspect is not very informative. Instead, it is more relevant to measure how well laypeople comprehend the text. \\
\textbf{Clarity:} Instead of asking evaluators to provide a single score for the simplicity \cite{jiang2020neural}, we evaluate their understanding by 
devising a set of questions to measure the usefulness of simplifications. The critical objective of radiology report simplification is to improve the clarity about the severity of the described conditions. There are two important dimensions of clarity: how well people understand the text and how well they believe they understand. Different combinations of those two dimensions can have different consequences for patients. For example, being confident while misinterpreting the text might lead to being too concerned or relaxed. Uncertainty is a clear indication that simplification was not adequate. Our survey is sequenced as in Figure \ref{fig:sur_lay}. We ask laypeople if they think they understand the original text (4 levels). Then, we ask them specifically if they think they understand the severity of the described condition (3 levels). This is followed by asking them to guess the severity, according to 5 severity levels defined in Appendix \ref{sec:severity}. 
This allows us to compare with the actual severity provided by a radiologist. We repeat those questions by suplementing the original sentence with simplification. Finally, we also ask them about their subjective opinion about the helpfulness of the simplification. 


We considered other ways to measure how well laypeople understand the text, such as quizzing them about the body parts and the meaning of the findings. We decided against it because it would be cumbersome to consistently convert the responses into numbers given a wide variety of radiology sentences. Also, asking this question would compound health literacy and simplicity. For example, even if a patient cannot fully understand the medical meaning, it could still be essential to hear that a condition impacting some part of their abdomen is not critical but requires a follow-up.  


\textbf{Preferences:}
Inspired by the design for evaluating text summarization \cite{goyal2022news}, we also ask evaluators to choose the most and the least preferred simplifications among multiple choices. In addition, layperson evaluators are encouraged to provide justifications for their selections, as shown in the right panel of Figure \ref{fig:sur_lay}. This free-text response can be used in qualitative analysis of laypeople's simplification preferences.

\section{Prompting and Self-Correction}
\subsection{Prompting ChatGPT}
\label{subsec:prompt}
Our preliminary results showed that ChatGPT can provide good simplifications of radiology sentences. Since we did not have a sufficiently large corpus of parallel text for radiology report simplification, we opted to use prompting without any fine tuning. Due to costs, we used ChatGPT-3.5 for all experiments in our study.

Prompt selection is partly an art form, so it was beyond the scope of this paper to comprehensively search for the best prompt for this application. Instead, we constructed two representative prompts after some trial and error -- one very simple (Plain) and another that relies on the Chain-of-thought (CoT) strategy, which makes ChatGPT think aloud while generating a response. All designed prompts can be found in Appendix \ref{subsubsec:generator}.

\subsection{Self-Correction Mechanism}
\label{sc}
\begin{figure}[h]
\includegraphics[width=0.98\linewidth]{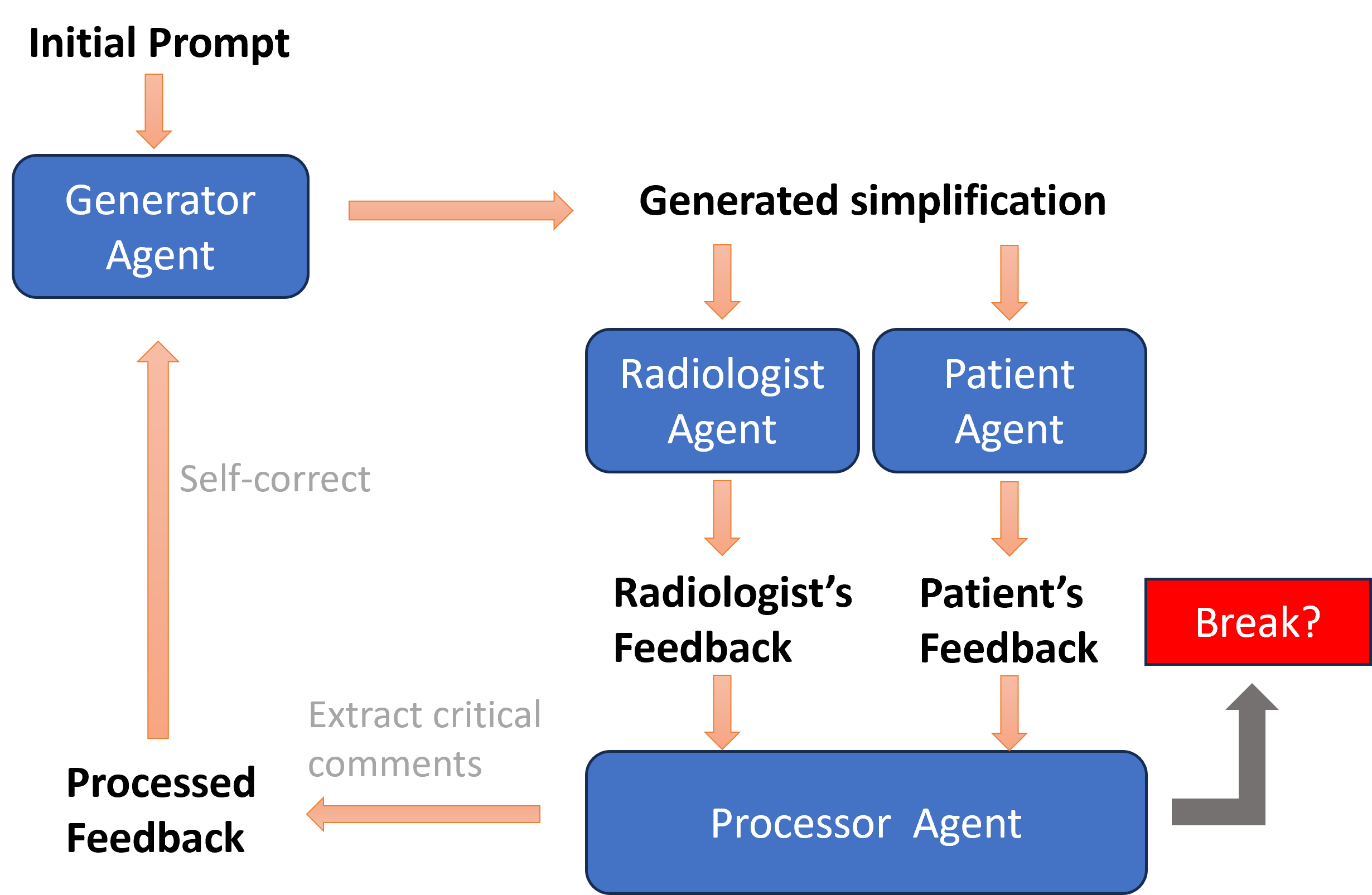}
\caption{The workflow of self-correction mechanism. Processor agent decides when to stop the process.}\label{fig:self}
\end{figure}

Inspired by \cite{madaan2023self}, we devised a Self-Correction mechanism for radiology report simplification. It relies on four differently instantiated ChatGPT agents: \textbf{Generator}, \textbf{Radiologist}, \textbf{Patient}, and \textbf{Processor}. The proposed workflow is shown in Figure \ref{fig:self}. Given an original radiology sentence, Generator is asked to generate a simplification. Then, Radiologist and Patient provide feedback about the simplification. Finally, Processor summarizes the feedback and provides the summary to Generator who is asked to improve the simplifications. This process iterates among these four agents until Processor determines that no further improvement is needed. This self-correction mechanism can be applied to LLMs without any model training.  

Inspired by \cite{park2023generative}, we instantiated Radiologist and Patient agents as distinct personas through distinct initial prompts shown in Appendix \ref{subsec:persona}. On the other hand, Generator and Processor agents are not prompted to become personas and are asked to provide an objective output. They are initialized using prompts that specify the task. Generator keeps the memory of conversation since it needs to refine the simplification based on the feedback from other agents. Generator is first provided a simple prompt for simplification. 
Feedback generated by Radiologist and Patient agents is summarized by Processor to reduce the redundancy. We asked Processor agent to first decide if there is any critical comment or improvement suggestion in the generated feedback. If so, Processor summarizes the feedback and passes it back to Generator using a 'refine prompt'. Otherwise, Processor generates a string starting with "No". In this case, the last simplification is saved as the self-correct simplification. The prompts are shown in Appendix \ref{subsubsec:processor}.

The proposed variant of self-correction mechanism is designed to imitate a conversation that could occur between a real radiologist and a patient to generate a good simplification of a radiology report. 
\begin{table*}[h] 
\tiny
\centering
\caption{FKGL, GFI, ARI scores and human evaluation results. In laypeople's evaluation, Q1: You understand the sentence? Q2: Can you guess the severity? Q3: What is the severity? Q4: Does simplification help you? Categorical answers are mapped to numeric types. Mean squared error (MSE) and accuracy (ACC) are presented for Q3.}
\resizebox{0.8\textwidth}{!}{%
\begin{tabular}{lccccc}\hline
Metrics &Original Sentence& Plain\_BS& Plain\_SC & CoT\_BS &CoT\_SC\\ \hline
\hline

FKGL $\downarrow$ &12.344 &8.813 &7.010&7.178&8.548\\
GFI $\downarrow$ &19.011 &14.632 &11.942 & 10.990&12.371\\
ARI $\downarrow$ &9.940 &6.463 &4.941 &4.404 &6.006\\

\hline
\hline

\multicolumn{6}{c}{Radiologist's Evaluation}\\
\hline
Correctness& 5.000&4.725& 4.650& 4.500&4.625\\
Completeness& 5.000&4.900& 4.675&4.775&4.875\\
Hallucination& 5.000&4.925&4.900 &4.850&4.825\\
Structure& 5.000&4.850&4.900 &4.825&4.875\\
Simplicity & 1.500&3.100&4.200 &4.375&4.575\\
\hline
\hline
\multicolumn{6}{c}{Laypeople's Evaluation}\\
\hline
Q1 (1 to 4) &1.801& 2.475 &3.225&3.341&\textbf{3.602}\\
Q2 (1 to 3) &1.579&1.825&2.325&2.398&\textbf{2.534}\\
Q3 (MSE $\downarrow$) &1.699&1.650&1.188&1.341&\textbf{1.068} \\
Q3 (ACC) &38.4\%&38.8\%&38.8\%&42.0\%&\textbf{52.3\%}\\
Q4 (-1 to 2) &N/A&0.613 &1.288&1.477&\textbf{1.705}\\
\hline
\end{tabular}}
\label{tab:res}
\end{table*}
\section{Experimental Design}
\subsection{Data}
For our experimental evaluation, we  identified 40 diverse, representative sentences from the radiology reports in the public database MIMIC III \cite{johnson2016mimic}. To arrive at those 40 sentences, a radiologist read 100 randomly selected MIMIC III radiology reports and marked
self-contained sentences about findings. The list of marked sentences was then narrowed down by removing redundancy and trying to maintain a diversity of findings in the reports. The selected sentences include a variety of findings (lesions, masses, obstructions, nodular surface, infection), conditions (being enlarged, abnormal, shrunken) of many anatomical parts in the abdomen (liver, kidney, pancreas, intestine, bones). Sentences
were selected to range from relatively simple to relatively complex. Attention was paid to ensuring that the chosen sentences were self-contained and did not require reading the surrounding sentences


\subsection{Types of Simplifications}
\label{type}

For each of the 40 radiology sentences, we produced four simplifications using ChatGPT-3.5. The first is Plain\_BS, which uses the plain prompt, while the second is CoT\_BS, which uses the CoT prompt, both introduced in Section \ref{subsec:prompt}. The remaining two use self-correction explained in Section \ref{sc}. The initial Generator prompt in self-corrected Plain\_SC is the plain prompt, while it is CoT prompt in CoT\_SC.
We used the same default temperature value for ChatGPT of 0.8 in all generations. 


\subsection{Automated Metrics}
In our study we did not generate ground truth simplifications because there might be a variety of acceptable simplifications with different lengths and levels of detail. As a result, our automated metrics only include three reference-free measures of simplicity: Flesch-Kincaid Grade Level (FKGL), Gunning Fog Index (GFI), and Automated Readability Index (ARI). The FKGL gives a grade-level score, indicating the minimum education level needed to comprehend the text. The GFI estimates the years of formal education required for understanding by focusing on sentence length and the frequency of complex words. The ARI, on the other hand, calculates readability based on the number of characters per word and words per sentence.

\subsection{Human Evaluation Protocol}
As described in Section \ref{eval}, we used two types of human evaluators to assess the quality of simplifications. 

\textbf{Radiologists.} We recruited one radiologist to evaluate the factuality of all simplifications as described in Section \ref{subsec:fact}. For further analysis, we also asked them to evaluate the simplicity via the question, "\textit{Do you think laypeople can understand the sentence?}". Likert scores in the range 1-5 were used for all questions. In addition, the radiologist was encouraged to provide justifications for the ratings. Moreover, we asked the radiologist to estimate the severity of described medical condition in each sentence using the five levels of severity as described in the Appendix \ref{sec:severity}. The distribution of severity scores on our data is: Critical: 4, Serious: 2, Moderate: 7, Mild: 20, Healthy: 7. The severity question is the same as Q3 in the survey for laypeople. This allowed us to evaluate the accuracy of laypeople's guesses of severity. 

\textbf{Laypeople.} We recruited eight laypeople to assess if the simplifications improve understanding. The participants were a mix of undergraduate and graduate students from a computer science department, none of whom had any training in medicine. Thus, they are representatives of highly-educated laypeople. For each of the 40 sentences, each layperson was given the questionnaire in Figure \ref{fig:sur_lay}. First, they answered 3 questions about the original sentence (left panel). Then, we selected one of the four simplified sentences using the Latin square design and asked them 4 questions from the middle panel to see if it improved their understanding compared to the original sentence alone. As the layperson was already starting to understand the original sentence, we did not repeat the middle panel questions for the 3 remaining simplifications. Instead, we asked a layperson the right panel questions to find which simplifications they liked the most and least. The simplifications in the right panel were listed at random to prevent bias.

Selected sentences, generated simplifications, and evaluation answers from the radiologist and laypeople are released.\footnote{https://github.com/Ziyu-Yang/Human-Evaluation}



\section{Results}
\subsection{Human Evaluation Results}

Top half of Table \ref{tab:res} shows results from the evaluation conducted by a radiologist. The 'Original Sentence' column denotes the scores assigned to the original radiology sentences. The factuality of the original radiology sentences was rated as five, by default.  Simplicity score for the original sentences was very low (1.50), indicating that most of the original sentences are not expected to be understood by laypeople. Simplicity score was much larger for simplified sentences and was the largest for the self-correction with CoT (CoT\_SC) approach. This result is consistent with the automated readability scores in the first three rows of Table \ref{tab:res}. It can be seen that readability of all 4 simplifications is significantly smaller (freshmen high school level) than for the original sentences (college level).


Factuality scores for all four types of simplifications remained close to  perfect. Hallucination and Structure scores were particularly high. Correctness scores were comparably lower, indicating occasional lack of precision in simplifications. Interestingly, factuality scores of Plain\_BS are higher than for the other three simplification methods. This reflects the trade-off between simplicity and factuality. We consider CoT\_SC the best approach because it achieved the highest simplicity with a very marginal decrease in factuality. 

\subsection{Do Simplifications Help?}
In the bottom half of Table \ref{tab:res}, we evaluate laypeople responses about simplicity. Q1, Q2, and Q3 in Figure \ref{fig:sur_lay} were designed to assess laypeople understanding of both the original sentences and their simplified versions. Q4 directly evaluated the effectiveness of these simplifications. We converted the categorical responses into numerical values\footnote{Q1: 'Not at all' -> 1; 'Completely' -> 4. Q2: 'Not at all' -> 1; 'High confidence' -> 3. Q3: 'Critical' -> 1; 'Healthy' -> 5. Q4: 'Furthered confused' -> -1; 'Much better' -> 2}. For the responses to Q3, we compared the participants' severity level choices with those of the radiologist and computed the Mean Squared Error (MSE) and Accuracy (ACC).

All simplifications had significantly higher simplicity scores than the original sentences on all questions. Notably, CoT\_SC achieved the highest scores accross all simplicity questions which is consistent with the radiologist's rating. 

\begin{table}[h]
\small
\caption{Confidence levels vs Mean squared errors and Accuracy for Q3 
}
\resizebox{.45\textwidth}{!}{%
\begin{tabular}{lccc}\hline
 & Not at all  & Low confidence & High confidence\\ \hline
MSE & 1.920 & 1.380 & \textbf{0.930 }\\
Accuracy &30.7\% & 39.8\% & \textbf{55.5\%} \\
\hline
\end{tabular}}
\label{tab:q3}
\end{table}
\subsection{Confidence vs Accuracy}
Table \ref{tab:q3} compares the correlation between the laypeople's confidence and the actual understanding of the severity of described medical conditions. When laypeople report the lowest confidence (Not at all), they also achieve the lowest accuracy in predicting severity (30.7\%), and when they report the highest confidence, the accuracy is the largest (55.5\%). However, there is still a significant gap between confidence and actual understanding. Even when highly confident, laypeople could correctly predict severity in just over half (55.5\%) of the sentences. We conclude that the simplifications might need to state the severity level explicitly.

\begin{figure}[h]
  \includegraphics[width=0.48\textwidth]{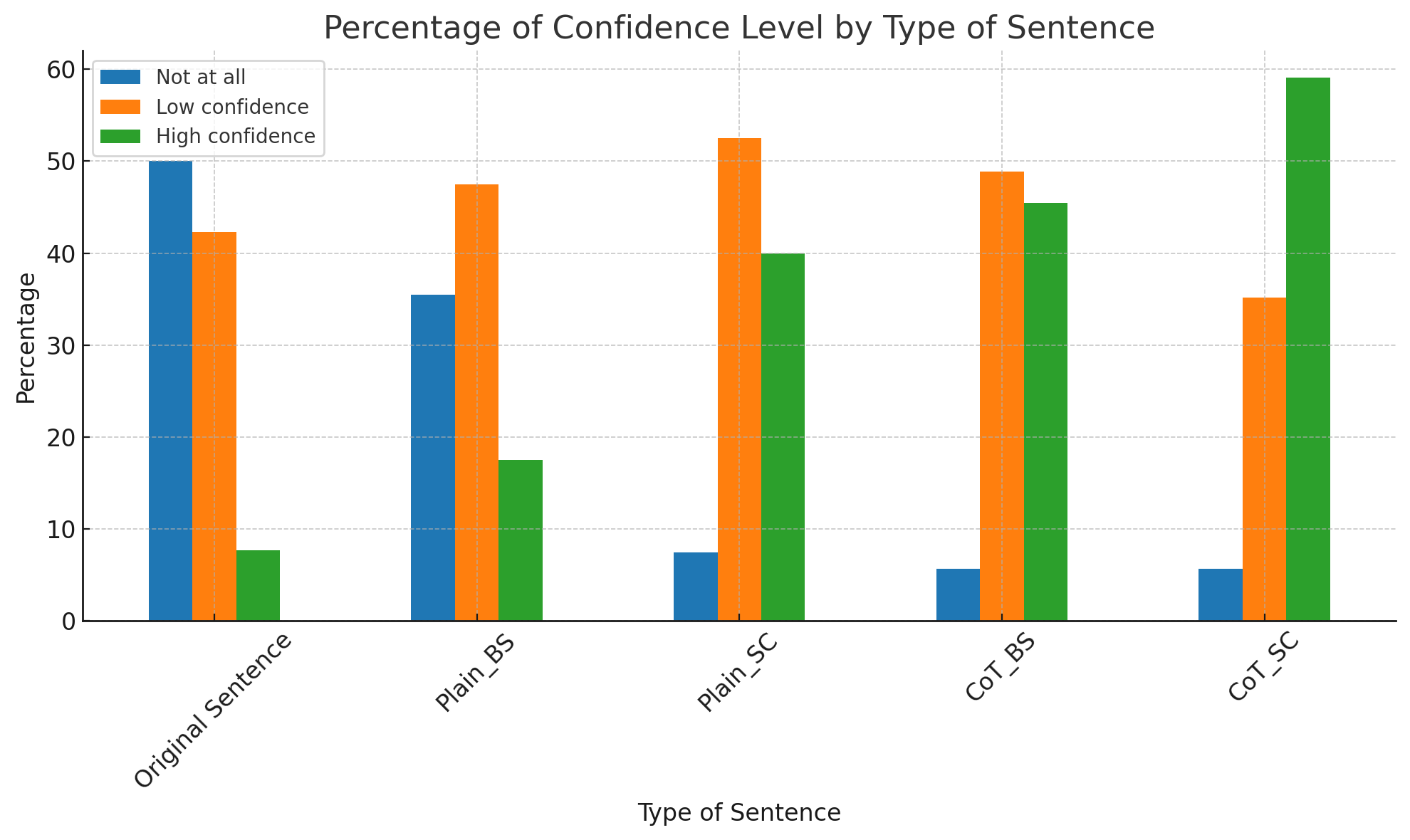}
  \caption{Distribution of confidence level (Q2) by laypeople given the original sentence and four types of simplifications}
  \label{fig:conf}
\end{figure}

To gain a deeper insight, in Figure \ref{fig:conf}, we show the distribution of confidence levels by laypeople for the original sentences and for each type of simplifications. We can see that all four types of simplifications are helpful, with CoT\_SC being the most successful.

\subsection{Which Simplifications are Preferred by Laypeople?}

In this subsection, we report on the preferences of laypeople towards different types of simplifications (right panel in Figure \ref{fig:sur_lay}). The findings are shown in Table \ref{tab:vote}, illustrating how often a specific simplification was deemed the most or least preferred based on the majority vote by the eight participants. The CoT\_SC simplification was the clear favorite compared to the other three variants. On the other hand, Plain\_BS simplification was the least favorite.

\begin{table}[h]
\small
\caption{Majority votes for the most and least preferences for all 40 sentences. 
}
\resizebox{.45\textwidth}{!}{%
\begin{tabular}{lcccc}\hline
 & Plain\_BS   & Plain\_SC & CoT\_BS & CoT\_SC\\ \hline
Most$\uparrow$ & 2 & 7& 15& \textbf{27} \\
Least$\downarrow$ &32 &7 &5 &\textbf{2} \\
\hline
\end{tabular}}
\label{tab:vote}
\end{table}

\begin{figure}[h]
  \includegraphics[width=0.48\textwidth]{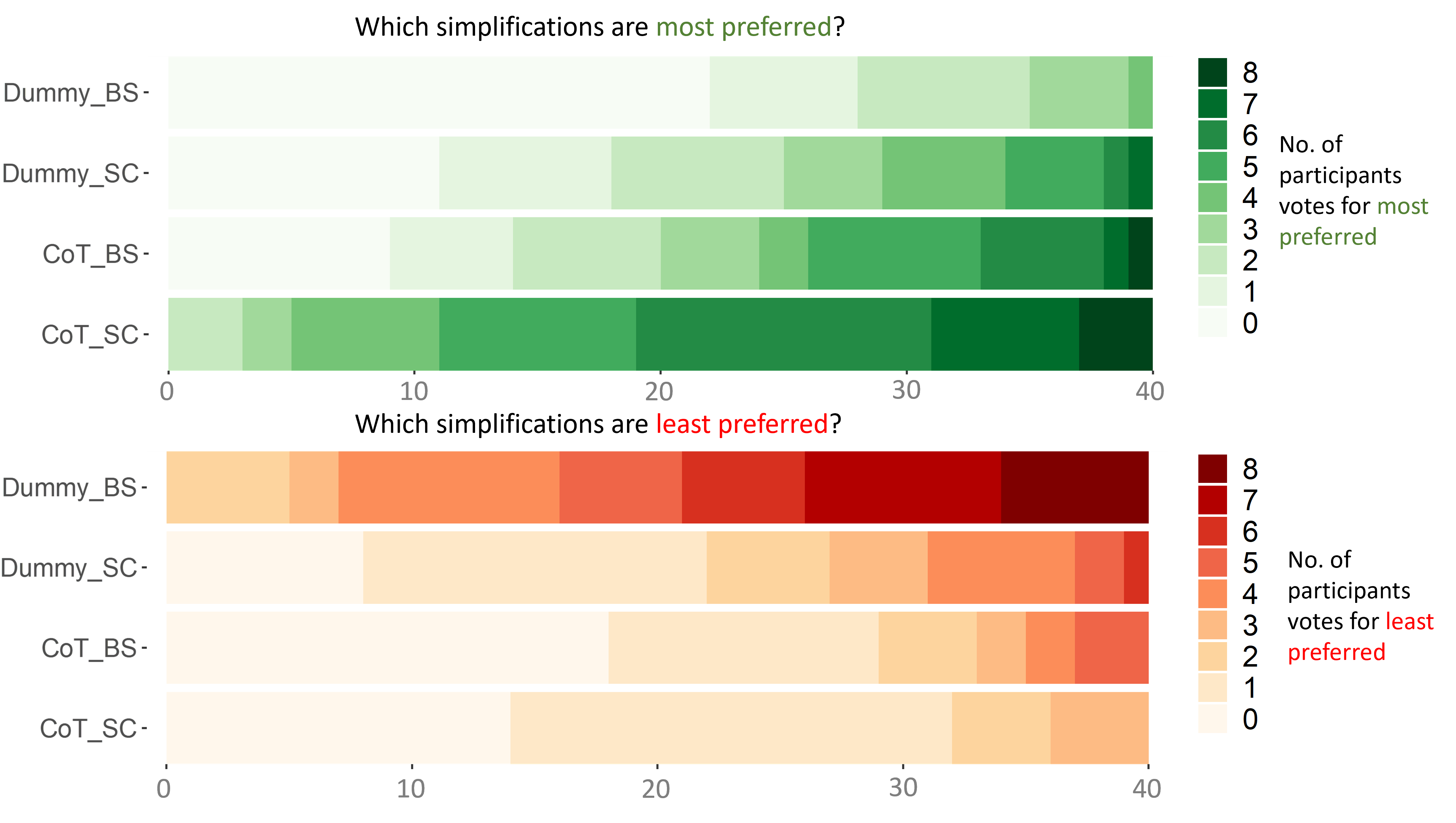}
  \caption{The horizontal stacked histogram of laypeople vote distribution for the most and least preferred simplifications.}
  \label{fig:vote}
\end{figure}
To further investigate laypeople's preferences, we adopted the analysis technique outlined in \cite{goyal2022news}. We calculated the inter-annotator agreement, applying Krippendorff's alpha with MASI distance \cite{passonneau2006measuring}, to account for the possibility of multiple selections for the best or worst simplifications in our research design. The alpha scores for the most and least preferred options were $0.177$ and $0.132$ respectively.

To visualize these results more clearly, we plot the distribution of laypeople's votes for each simplification version in Figure 4. The distribution of votes highlights that CoT\_SC was chosen as the most preferred simplification by at least half of the participants in over 90\% of the sentences. Additionally, both CoT\_BS and Plain\_SC received votes as the most preferred simplification from at least one participant for approximately 80\% of the sentences. Unsurprisingly, based on the vote distribution, Plain\_BS was the least preferred simplification version.


\subsection{Factual Errors in Generations}
ChatGPT is providing impressive simplifications, but it is not perfect. We observed that the slight decrease in factuality scores was caused by a few outliers. We noticed that during the self-correction process, Generator agent can occasionally generate incorrect information and hallucinate as it is trying to addresses feedback from Radiologist and Patient agents. There were in total eight factual errors observed by the radiologist evaluator among simplifications by the two self-correction approaches.
\begin{table}[h]
\scriptsize
\caption{Four representative examples where the radiologist pointed out factual errors in generated simplifications. The radiologist's comments are included.}
\begin{tabularx}{7.8cm}{SlV{2}X}
\hline
\hline
\textbf{ORI}:& {Few stable less than 5 mm hypodensities, too small to characterize, probably simple cysts.}\\
\textbf{SIM}: & {There are a few tiny fluid-filled sacs, called cysts, that show up as lighter areas on the scan. These cysts are very small, measuring less than 5 mm.} \\

\hline
\textbf{Comments}: & {It says there are cysts - the radiologist was not definitive. Original text said it is PROBABLY cysts. (\textcolor{red}{\textbf{Completeness error}})}      \\    

\hline
\hline
\textbf{ORI}:& {Atrophic kidneys are noted bilaterally with multiple simple cysts.}\\
\textbf{SIM}: & {The kidneys have become smaller and are not functioning well. There are fluid-filled sacs in both kidneys.} \\

\hline
\textbf{Comments}: & {You cannot assess whether they functional well based on CT. (\textcolor{red}{\textbf{Hallucination error})}}     \\    
\hline
\hline
\textbf{ORI}:& {Status post hysterectomy.}\\
\textbf{SIM}: & {After having a hysterectomy, which is a surgery to remove the uterus, the person is in the recovery phase.} \\

\hline
\textbf{Comments}: & {The sentence simply says surgery was done sometime in the past - not that the patient is in recovery. (\textcolor{red}{\textbf{Hallucination error}})}     \\    
\hline
\hline
\textbf{ORI}:& {Small calcified granulomata in the left hepatic lobe.}\\
\textbf{SIM}: & {There are small hardened clusters of immune cells in the left part of the liver.} \\

\hline
\textbf{Comments}: & {They are not immune cells. (\textcolor{red}{\textbf{Correctness error}})}     \\

\hline
\hline
\end{tabularx}
\label{tab:vis_rad}
\end{table}
In Table \ref{tab:vis_rad}, we show and analyze four examples to point to typical errors. ChatGPT may generate hallucinations through over-interpreting the sentences. We can also see that ChatGPT may not use perfectly precise simple words to explain the complex medical jargon. However, ChatGPT is overall very impressive in producing high-quality text simplifications and is quite capable of self-correction.

\begin{table}[h]
\scriptsize
\caption{Representative examples where laypeople voted CoT\_SC as their most preferred simplification. Laypeople justifications are listed. The ground-truth severity level is appended to the end of the original sentences. The laypeople's comments are included.}
\begin{tabularx}{7.8cm}{SlV{2}X}
\hline
\hline
\textbf{ORI}:& {There are metastatic bony lesions throughout the thorax. (\textcolor{blue}{Critical})}\\
\textbf{CoT\_SC}: & {The doctors found that cancer from somewhere else has spread to the bones in the chest.	} \\

\hline
\textbf{Comments}: & {It explains well.}\\

\hline
\hline
\textbf{ORI}:& {There is grade 1 anterolisthesis of L4 on L5. (\textcolor{blue}{Mild})}\\
\textbf{CoT\_SC}: & {The diagnosis indicates a \textbf{mild} condition where the fourth bone in the lower part of the spine is slipping slightly forward over the fifth bone. This finding suggests a relatively \textbf{low level of severity}.} \\

\hline
\textbf{Comments-1}: & {The language is simple and easy to understand. I like that it tells the severity of the condition.}\\
\textbf{Comments-2}: & {Accurate and detailed.}\\
\hline
\hline
\textbf{ORI}:& {There is no intrahepatic or extra-hepatic biliary ductal dilatation. (\textcolor{blue}{Healthy})}\\
\textbf{CoT\_SC}: & {The tubes both inside and outside the liver that carry bile (a digestive fluid) are not enlarged or widened, which is \textbf{usually a good sign} and means there is likely no blockage.} \\

\hline
\textbf{Comments-1}: & {I like that it mentions "usually a good sign" because it indicates the severity of the condition.}\\
\hline
\hline
\textbf{ORI}:& {Partially visualized central pulmonary arteries are not dilated. (\textcolor{blue}{Healthy})}\\
\textbf{CoT\_SC}: & {The part of the blood vessels in the middle of the lungs that we can see is not enlarged, which is \textbf{normal}.} \\

\hline
\textbf{Comments-1}: & {Actually tells me that it's normal.}\\
\textbf{Comments-2}: & {Normal was important to me.}\\
\hline
\hline
\textbf{ORI}:& {The osseous structures are diffusely demineralized. (\textcolor{blue}{Moderate})}\\
\textbf{CoT\_SC}: & {The bones throughout the body have lost minerals and become weaker. This widespread loss of minerals may have implications for the overall strength and health of the bones.} \\

\hline
\textbf{Comments}: & {Straight to the point. I needed to know that the bones are weakened.}\\

\hline
\hline
\end{tabularx}
\label{tab:vis_pat}
\end{table}

\subsection{Preferences of Laypeople}
In Table \ref{tab:vis_pat}, we show five examples from laypeople responses and discuss why CoT\_SC could enhance comprehension. These examples are selected because they are representatives of sentences with different severity levels. We observe that participants prefer a simplification that 1) explains the medical condition in detail, 2) uses simple language, 3) indicates the severity of the condition. The CoT\_SC simplification in the second example implies a mild severity level, which is not explicitly stated in the original sentence.

\section{Conclusion}
This paper introduces a two-pronged approach for human evaluation of radiology report simplifications. It proposes a specialized variant of the self-correction mechanism that allows ChatGPT to generate high-quality simplifications. The analysis of results derived from human evaluation show that our proposed evaluation protocol successfully reveals diverse facets of simplification quality.






\section{Limitations}
The first limitation of our study is that it focuses on simplification of individual sentences. Descriptions of some radiology findings are complex and require multiple sentences. While we do not expect LLMs to struggle with simplifying multiple sentences, an additional challenge would be extracting multi-sentence findings. 

The second limitation is associated with simplifying the whole reports that often have multiple findings. While a trivial approach might consist of chunking the text into logical units and simplifying each unit separately, this approach might result in overly long simplification. Thus, it might be necessary to identify and simplify only the most significant findings from the report.  

The third limitation is that we used only 40 original radiology sentences in the experimental evaluation. Ideally, we would like to consider a much larger set of sentences. However, the cost associated with this would be prohibitive. There are large computational costs associated with the self-correcting algorithms because they require multiple calls to ChatGPT to create a single simplification. There are also significant costs associated with human evaluation. It took laypeople over two hours on average to finish all the needed evaluations. It took the radiologist even longer. We estimated that 40 sentences were the minimum that allowed us to evaluate our ideas. We note that we made an effort to make those sentences representative of the radiology report diversity.

The fourth limitation of the study is that we obtained expert evaluation from a single radiologist. In fact, we recruited two more volunteer radiologists for our research, but neither was able to to finish the evaluation due to its length. Thus, we decided not to use their partial responses in the paper. It will be important for future studies to recruit multiple radiologists to estimate the factualness better and obtain a more complete understanding of their simplification preferences. It would also allow us to measure the inter-rater reliability. To be more successful, we  need to make our survey easier to complete.

The fifth limitation is that our laypeople were college-educated individuals. It would be important in future research to recruit a more diverse group of laypeople and paint a more complete picture of the quality of simplifications and preferred types of simplifications. 

\bibliography{acl_latex.bib}

\appendix
\clearpage
\section{Appendix}
\label{sec:appendix}
\subsection{Designed Survey for Radiologist}
\label{subsec:rad}
\begin{figure*}[h]
  \includegraphics[width=\textwidth]{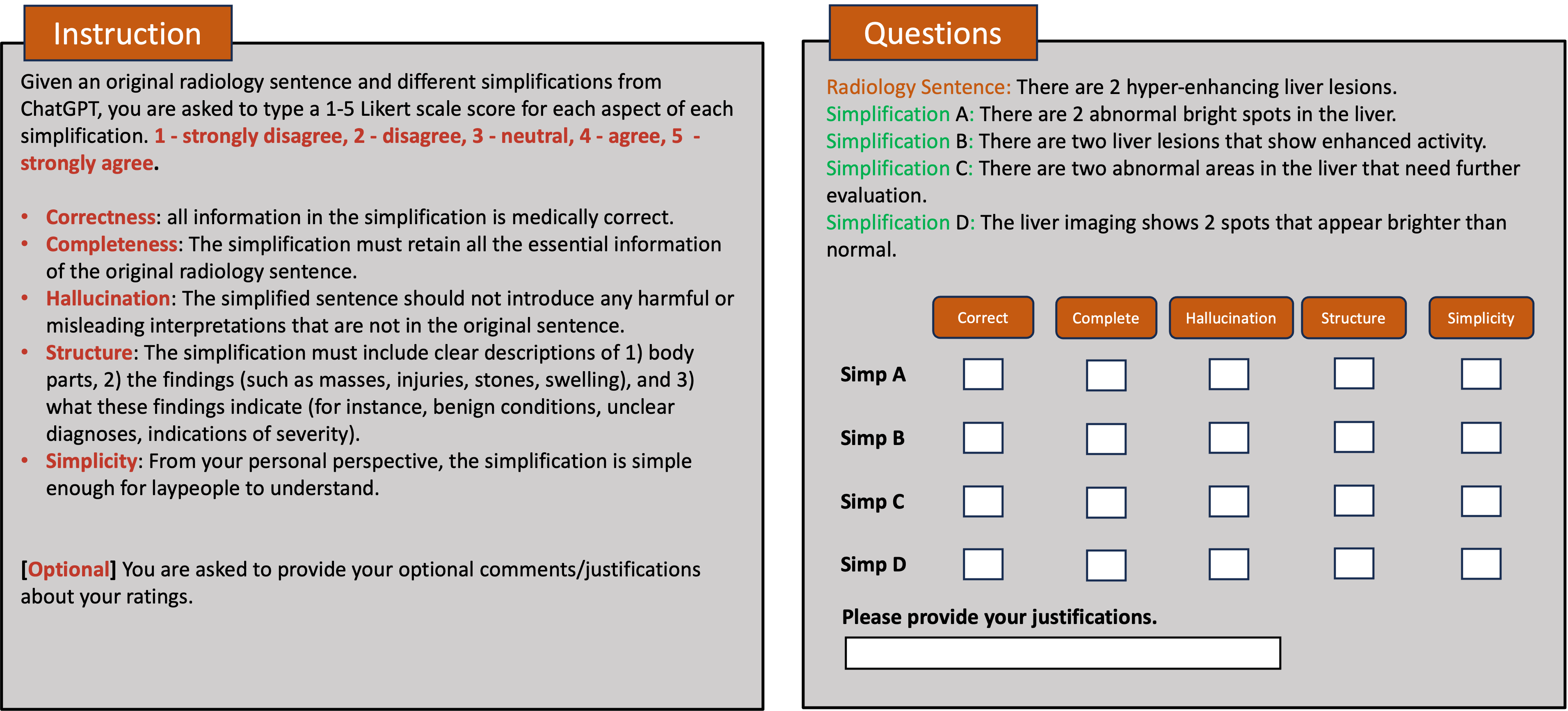}
  \caption{Expert evaluation of radiology report simplification. (\textbf{left panel}) lists instructions, (\textbf{left panel}) is a survey form with text boxes for ratings and justification.
}
  \label{fig:sur_rad}
\end{figure*}

The exact design of the survey for the radiologist is shown in Figure \ref{fig:sur_rad}
\subsection{Definitions of Severity Levels}
\label{sec:severity}
\begin{itemize}
    \item CRITICAL (5): Describes a medical condition that poses a threat to a person's life. A critical condition requires urgent care and close monitoring. 
    \item SERIOUS (4): Describes a condition that requires medical attention but is not immediately life-threatening. Treatment may involve hospitalization, medication, or other interventions. 
    \item MODERATE (3): Describes a condition that is not severe but may require medical attention and treatment. The condition may cause discomfort or affect a person’s ability to carry out normal activities. 
    \item MILD (2): Describes a condition that is not serious. The condition may cause minor discomfort or inconvenience but is unlikely to have a significant impact on a person’s overall health. 
    \item HEALTHY (1): Findings that are considered normal or benign with no significant abnormalities.
\end{itemize}

\subsection{Prompts}
\label{subsec:b}

This section provides details about the design of four ChatGPT agents used in the self-correction mechanism outlined in Figure 2. 

\subsubsection{Generator Agent}
\label{subsubsec:generator}
Generator is initialized with a simple objective prompt. We considered two specific prompts as follows:

\begin{itemize}
    \item Plain prompt:\\ \textit{Simplify the sentence: <RADIOLOGY SENTENCE>. }
    \item CoT prompt:\\ \textit{Sentence: <RADIOLOGY SENTENCE>. \\Can you list all the complicated medical terms and provide explanations that are understandable by laypeople? Finally, write a simplification of the original sentence that laypeople can understand. }
\end{itemize}
The response from Generator is saved as evaluated in our experiments. In addition, the response from Genarator is used to start the self-correction mechanism illustrated in Figure 2.

\subsubsection{Radiologist Agent} Human radiologists can adequately evaluate the factualness of radiology report simplifications. We mimic this by creating a Radiologist agent with an initial prompt that ask ChatGPT to pretend to have a persona of radiologist, following the related idea presented in  \cite{park2023generative, li2023camel}. 

Text in blue in Figure \ref{fig:rad} defines Radiologist persona. Text in green is an instruction consistent with the survey we designed for human radiologists and that was used in human evaluation of simplifications.

\subsubsection{Patient Agent} Similar to Radiologist agent, we created a Patient agent to provide feedback about the understandability of the simplification from Generator. As shown in Figure \ref{fig:pat}, we asked Patient to act as a layperson who lacks medical knowledge and cannot understand complex medical concepts. Further instructions and warnings are specified to avoid generating comments that are beyond the ability of a layperson. 


\subsubsection{Processor Agent} 
\label{subsubsec:processor}
The feedback generated by Radiologist and Patient agents is summarized by Processor to reduce the redundancy. We asked Processor agent to first decide if there is any critical comment or improvement suggestion in the generated feedback. If so, Processor summarizes the feedback and passes it back to Generator using a 'refine prompt'. Otherwise, Processor generates a string starting with "No". In this case, the last simplification is saved as the self-correct simplification. The following is a prompt we used for Processor:
\begin{itemize}
    \item \textbf{Initial prompt for Processor}:\\
    \textit{Feedback: <FEEDBACK>\\
    Are there any critical comments or improvement suggestions in Feedback? If so, extract them starting with "Yes". Otherwise, say "No".}
\end{itemize}
 The following prompt is used to ask Generator to improve its previous simplification:
\begin{itemize}
\item \textbf{Refine prompt for Generator}:\\
    \textit{Radiologist's feedback: <PROCESSED FEEDBACK>\\
    Patient's feedback: <PROCESSED FEEDBACK>\\
    Can you improve your simplification while keeping it concise?}
\end{itemize}
\subsection{Personas}
\label{subsec:persona}
\begin{figure}[h]
\includegraphics[width=\linewidth]{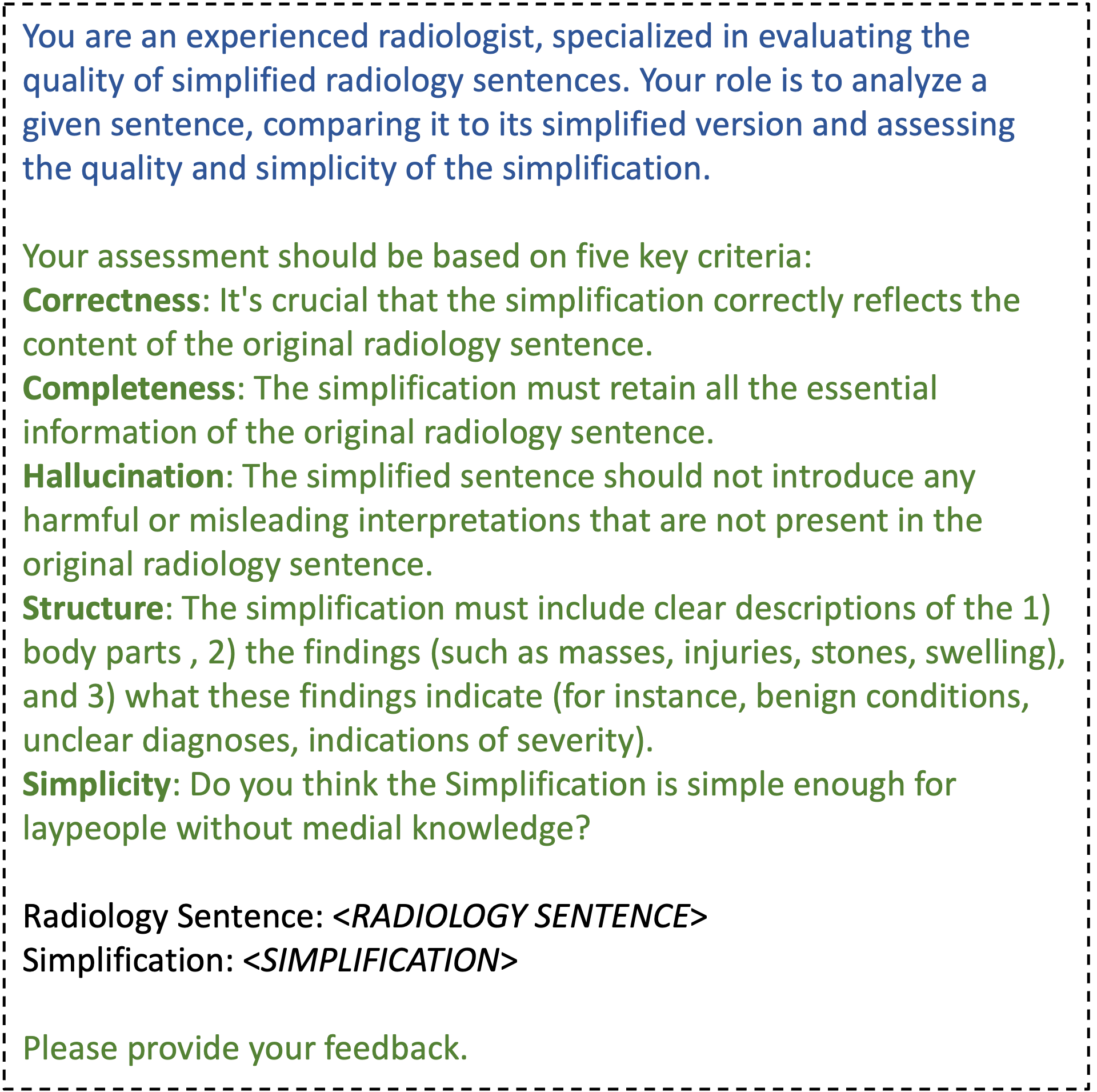}
\caption{The persona of \textbf{Radiologist} agent and task instructions}\label{fig:rad}
\end{figure}

\begin{figure}[h]
\includegraphics[width=\linewidth]{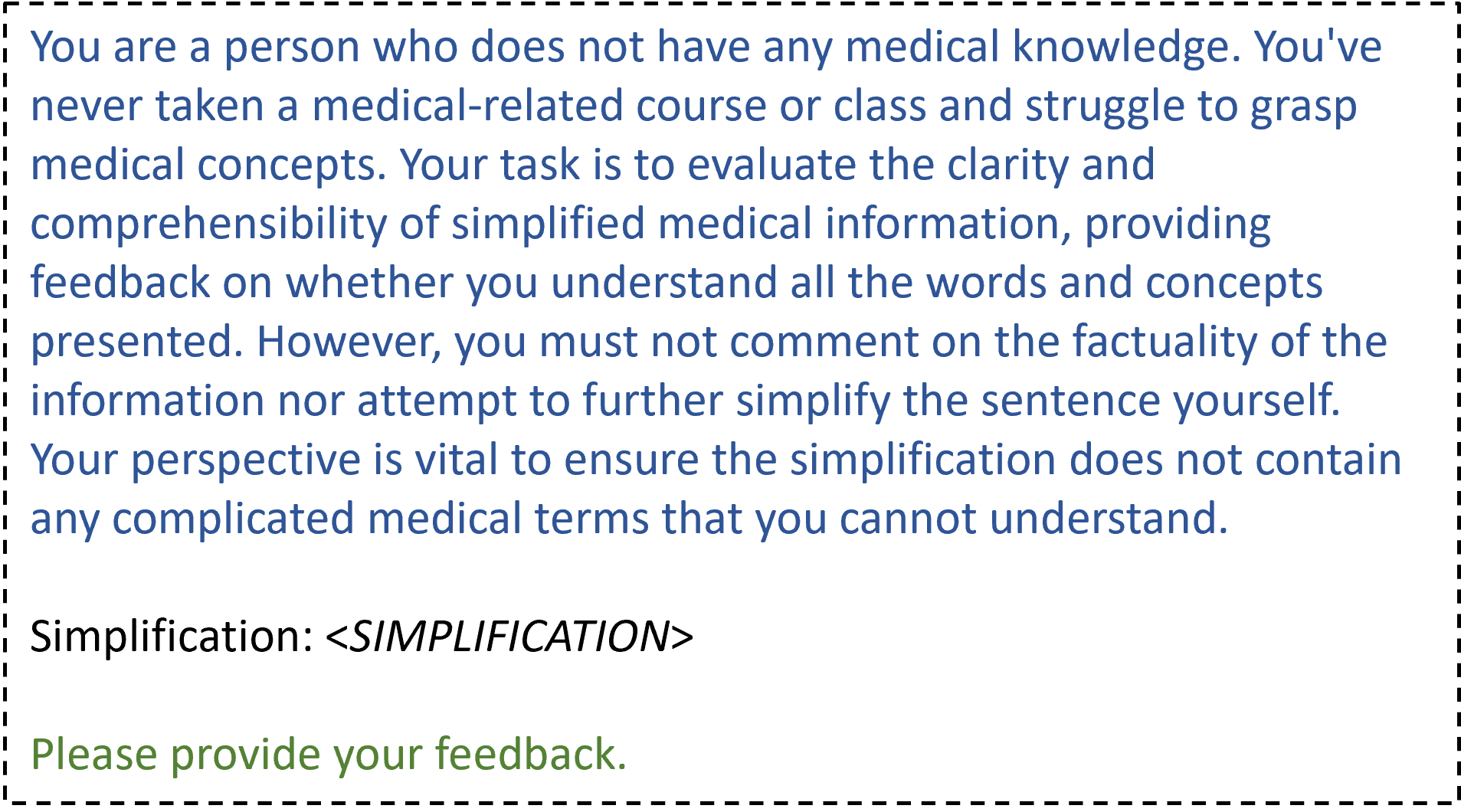}
\caption{The persona of \textbf{Patient} agent and task instructions}\label{fig:pat}
\end{figure}

The designed personas for the Radiologist and Patient agents are shown in Figure \ref{fig:rad} and \ref{fig:pat}
\\

\end{document}